\documentclass[11pt]{article}

\usepackage[margin=1in]{geometry}
\usepackage{times}
\usepackage{graphicx}
\usepackage{amsmath}
\usepackage{amssymb}
\usepackage{booktabs}
\usepackage{hyperref}
\usepackage{microtype}
\usepackage{xcolor}
\usepackage{caption}
\usepackage{subcaption}
\usepackage{float}
\usepackage{natbib}
\usepackage{placeins}

\hypersetup{
    colorlinks=true,
    linkcolor=blue,
    citecolor=blue,
    urlcolor=blue
}

\title{\textbf{Zero-Label Driving Scenario Complexity Detection\\
via Joint Embedding Predictive Architecture}}

\author{
  Santosh Jaiswal \\
  Independent Researcher \\
  \texttt{hellojais@gmail.com}
}

\date{}

\begin{document}

\maketitle

% ─────────────────────────────────────────────────────────────────
\begin{abstract}
Identifying complex and safety-critical driving scenarios in large
unlabelled datasets is an important but expensive problem. Existing
approaches rely on human annotators, supervised classifiers, or
carefully engineered rule sets, all of which require substantial
prior knowledge about what constitutes a difficult scenario.
We ask whether a model can discover scenario complexity on its own,
with no labels at any stage.
We train a minimal Joint Embedding Predictive Architecture (JEPA)
on structured agent state data from the nuPlan mini dataset and
use the temporal prediction error as a zero-shot complexity score.
Without access to any ground-truth labels during training or
evaluation setup, the model assigns significantly higher scores to
scenarios involving unprotected turns, crosswalk interactions, and
pedestrian proximity, and significantly lower scores to lane-following
and stationary-traffic scenarios.
We validate this finding through four ablation experiments that
isolate the source of the signal, and through a downstream anomaly
detection evaluation that achieves Average Precision of 0.512
against a 0.436 chance baseline.
The results show that temporal prediction error in a self-supervised
latent world model is a practical proxy for driving scenario complexity.
\end{abstract}

% ─────────────────────────────────────────────────────────────────
\section{Introduction}

Real-world autonomous vehicle datasets are heavily skewed toward
uneventful driving \citep{nuplan2023}. A typical recording session
contains far more straight-road cruising than unprotected left turns
or pedestrian crossings. This imbalance creates a practical problem:
to test or improve a planning system, engineers must manually find the
rare complex scenarios buried inside millions of routine clips.

Several approaches address this. Rule-based methods flag scenarios
by speed thresholds or proximity to road objects, but cannot capture
complex multi-agent interactions \citep{ding2025surprise}.
Supervised classifiers require labelled training data, which is
expensive to collect and biased by the label taxonomy. Human
preference collection, as used by \citet{ding2025surprise},
produces high-quality rankings but does not scale automatically
to new datasets.

We take a different approach. We train a world model that learns
to predict the future from the past, and treat the prediction
error as a complexity signal. Scenarios where the model is wrong
are by definition scenarios the model did not expect. Our hypothesis
is that the scenarios a trained model finds most surprising are
the same scenarios a human driver would find most demanding.

The technical foundation is the Joint Embedding Predictive
Architecture proposed by \citet{lecun2022jepa}. Rather than
predicting the future in raw input space, JEPA predicts a compact
latent representation of the future. This forces the model to capture
the meaningful structure of the scene rather than irrelevant details
such as sensor noise or background appearance.

We apply this framework to structured agent state data from nuPlan
\citep{nuplan2023}, where each agent is described by its position,
velocity, heading, and type. This representation is already semantic,
which means the latent space the model learns is organised around
agent dynamics and interactions rather than visual properties.

Our specific contributions are:

\begin{itemize}
  \item We demonstrate that a JEPA model trained on structured
        driving agent data with no labels produces surprise scores
        that correlate with ground-truth nuPlan scenario tags.
  \item We present four ablation experiments that confirm the signal
        comes specifically from the trained JEPA weights and from
        the EMA training mechanism.
  \item We show that constant-velocity kinematic baselines actively
        misrank complex scenarios because they assign low surprise
        to slow-approach events that precede complex manoeuvres.
  \item We provide a zero-shot anomaly detection evaluation achieving
        AP of 0.512 versus a 0.436 chance baseline, with consistent
        improvement across the full ranking curve.
\end{itemize}

% ─────────────────────────────────────────────────────────────────
\section{Related Work}

\subsection{Surprise and Interactivity Metrics in Autonomous Driving}

\citet{ding2025surprise} define surprise potential as the
distribution shift in predicted agent trajectories under
counterfactual perturbation. Given a scenario, one agent's
trajectory is modified and a motion predictor measures how
much other agents react. Large reactions indicate high
interactivity. This method requires three components: a
counterfactual generator, a trajectory prediction model,
and a distribution shift metric. Our method requires only
a single trained JEPA model and uses temporal prediction
error directly, without counterfactual generation. We also
validate against objective scenario tags rather than
human preference labels.

\subsection{JEPA and Self-Supervised World Models}

\citet{lecun2022jepa} proposed JEPA as an alternative to
generative self-supervised learning. The key insight is that
predicting in a learned latent space avoids the computational
cost and training instability of reconstructing irrelevant
details. I-JEPA \citep{assran2023ijepa} demonstrated this
on images and V-JEPA \citep{bardes2024vjepa} extended it to
video.

Concurrently, \citet{wang2026drivejepa} apply V-JEPA to raw
driving camera video as a pretraining step for trajectory
planning. Their goal is improved planning performance on
benchmark tasks. Our work differs in input modality
(structured agent states versus raw pixels), training scale
(1,322 short clips versus large-scale video), and objective
(unsupervised complexity scoring versus planning).

\subsection{Unsupervised Scenario Mining}

Prior work on scenario mining largely falls into supervised
and rule-based categories \citep{ding2025surprise}. Unsupervised
approaches include clustering over trajectory features
\citep{nuplan2023} and prediction error baselines. We are
not aware of prior work that applies JEPA-style temporal
prediction to structured agent dynamics specifically for
zero-shot complexity scoring.

% ─────────────────────────────────────────────────────────────────
\section{Method}

\subsection{Data Representation}

We use the nuPlan mini dataset \citep{nuplan2023}, which contains
64 SQLite files recording real urban driving. From these files
we extract 1,322 scenarios, each representing 5 seconds of
driving at 10~Hz. Each scenario is stored as a tensor of
shape $[50, 21, 6]$: 50 timesteps, 21 agents (one ego vehicle
plus up to 20 surrounding agents padded to a fixed size), and
6 features per agent.

The six features are position $(x, y)$, velocity $(v_x, v_y)$,
heading $\theta$, and a categorical agent type. Positions are
normalised by dividing by 50~m and velocities by 10~m/s, so that
typical values fall near $[-1, 1]$.

\subsection{Temporal Split}

Each scenario is split at the midpoint into a context window
(frames 0--24) and a target window (frames 25--49):

\begin{equation}
  \mathbf{X}_{\text{ctx}} = \mathbf{X}_{0:25}, \quad
  \mathbf{X}_{\text{tgt}} = \mathbf{X}_{25:50}
\end{equation}

The model receives only the context window as input and must
predict a representation of the target window. The target
window is never provided to the context encoder during inference.

\subsection{Architecture}

\begin{figure}[htbp]
  \centering
  \includegraphics[width=0.85\textwidth]{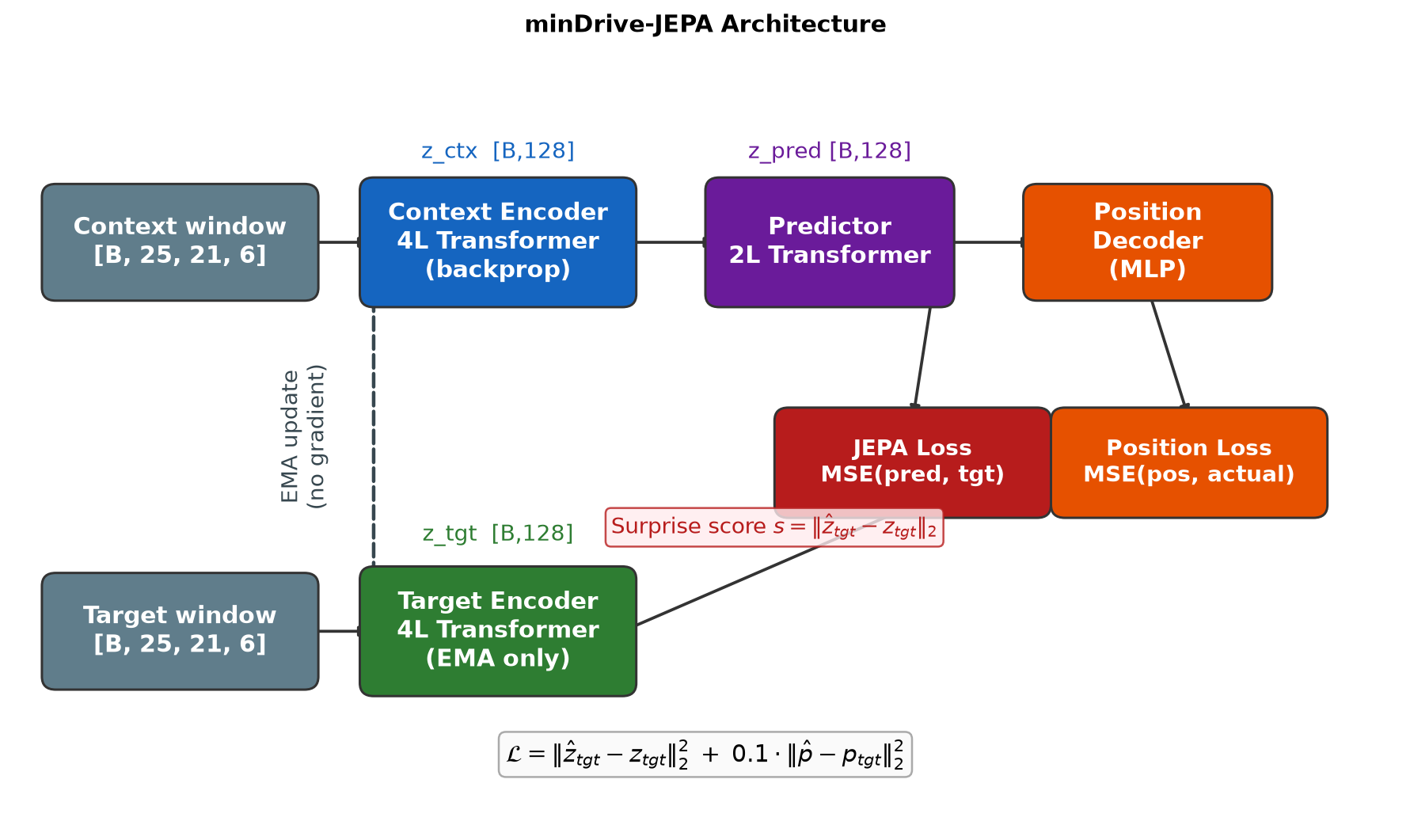}
  \caption{Overview of the minDrive-JEPA architecture. The context
  encoder and predictor are trained by gradient descent. The target
  encoder receives no gradients and is updated solely by EMA of the
  context encoder weights.}
  \label{fig:arch}
\end{figure}

The model has four components, illustrated in Figure~\ref{fig:arch}.

\textbf{Context encoder.} A Transformer encoder with 4 layers,
128-dimensional hidden state, 4 attention heads, and
feedforward dimension 512. The input is flattened to
$[B, T, 126]$ (21 agents times 6 features), linearly projected
to dimension 128, and augmented with sinusoidal positional
encodings. The output is mean-pooled across the time dimension
to produce a single context latent $\mathbf{z}_{\text{ctx}} \in \mathbb{R}^{128}$.

\textbf{Predictor.} A 2-layer Transformer encoder that maps
the context latent to a predicted future latent
$\hat{\mathbf{z}}_{\text{tgt}} \in \mathbb{R}^{128}$.
The predictor is given a learnable horizon embedding to condition
on prediction distance.

\textbf{Target encoder.} An identical copy of the context encoder
whose weights are updated only by exponential moving average (EMA)
and never by backpropagation. It reads the target window and
produces the target latent $\mathbf{z}_{\text{tgt}} \in \mathbb{R}^{128}$.

\textbf{Position decoder.} A small MLP that decodes the predicted
latent into predicted future positions $\hat{\mathbf{p}} \in \mathbb{R}^{21 \times 2}$,
used as an auxiliary supervision signal.

\subsection{Training Objective}

The total loss combines a JEPA latent prediction term and an
auxiliary position reconstruction term:

\begin{equation}
  \mathcal{L} = \underbrace{\|\hat{\mathbf{z}}_{\text{tgt}} - \mathbf{z}_{\text{tgt}}\|_2^2}_{\text{JEPA loss}}
  + \lambda \underbrace{\|\hat{\mathbf{p}} - \mathbf{p}_{\text{tgt}}\|_2^2}_{\text{position loss}}
\end{equation}

where $\lambda = 0.1$. Gradients flow through the context encoder
and predictor but are stopped at the target encoder boundary.

\textbf{EMA update.} After each training step the target encoder
weights $\theta_{\text{tgt}}$ are updated as:

\begin{equation}
  \theta_{\text{tgt}} \leftarrow \alpha \, \theta_{\text{tgt}} + (1 - \alpha) \, \theta_{\text{ctx}}
\end{equation}

with $\alpha = 0.996$. This keeps the target encoder as a smoothed
historical average of the context encoder, preventing the trivial
collapse where both encoders converge to a constant output.

\subsection{Surprise Score}

At inference, the surprise score for a scenario is the L2 distance
between the predicted and actual target latents:

\begin{equation}
  s = \|\hat{\mathbf{z}}_{\text{tgt}} - \mathbf{z}_{\text{tgt}}\|_2
\end{equation}

A high score indicates the model's prediction of the future was
far from the actual future in latent space. A low score indicates
the future was largely as expected. No labels are used at any
point.

\subsection{Training Details}

We train for 50 epochs with batch size 32, learning rate $3 \times 10^{-4}$
with linear warmup over 5 epochs followed by cosine decay, weight
decay $10^{-4}$, dropout 0.2, and gradient clipping at 1.0. An
80/20 train/validation split is used. The model has 1,289,130
trainable parameters. Best validation loss of 0.0345 is achieved
at epoch 12. All experiments use an Apple M5 Max GPU via the MPS
backend.

% ─────────────────────────────────────────────────────────────────
\section{Experiments}

\subsection{Tag Correlation}

nuPlan provides ground-truth \texttt{scenario\_tag} labels for each
scenario, with 67 distinct tag types covering a wide range of
behavioural categories (58 tags have $\geq$5 scenario occurrences
and are used in this analysis). These labels were not used during
training or in any way to define the surprise score.

\begin{figure}[htbp]
  \centering
  \includegraphics[width=\textwidth]{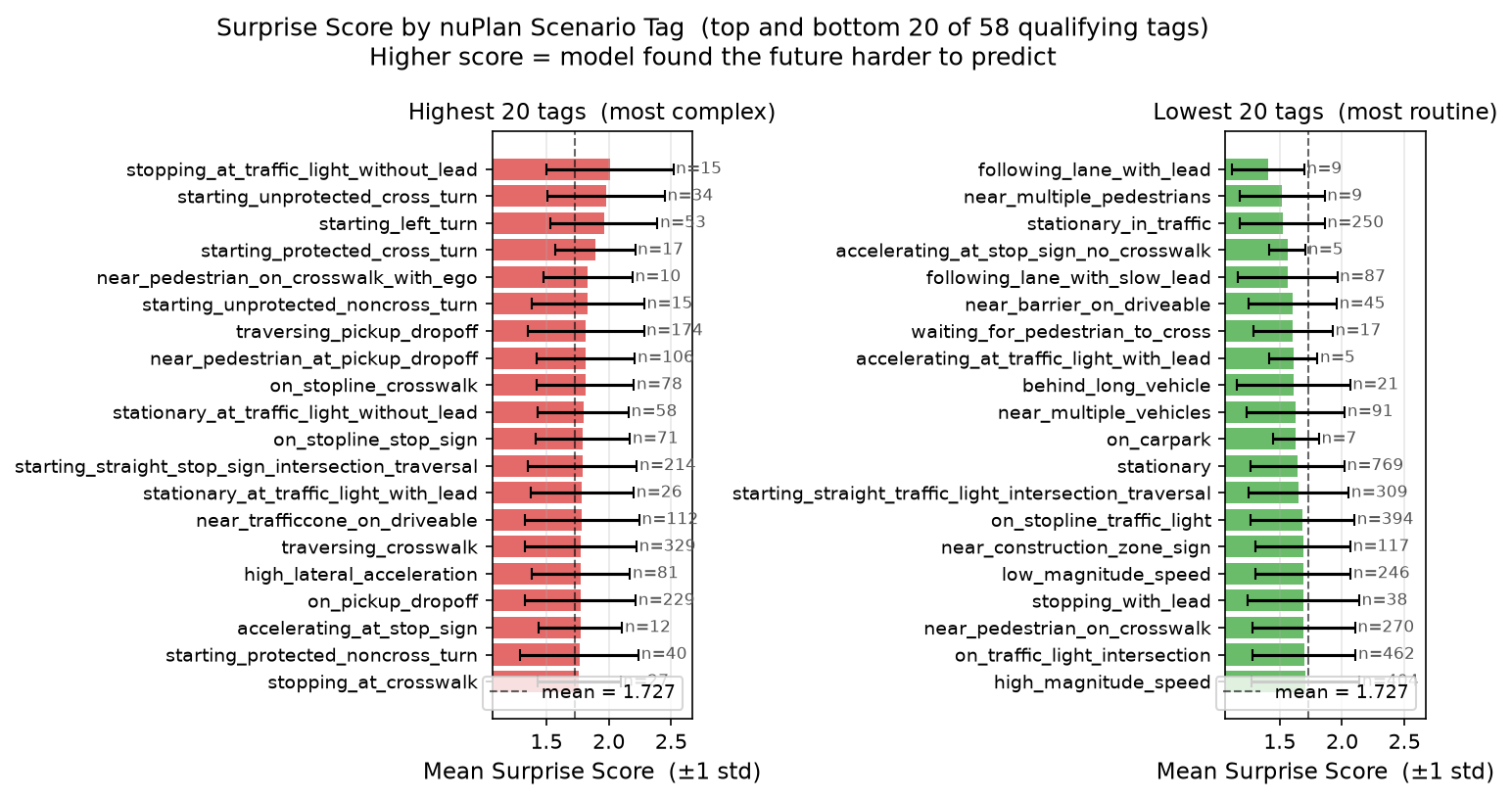}
  \caption{Mean surprise score per nuPlan scenario tag. Left panel:
  top 20 most surprising tags (complex manoeuvres, in red).
  Right panel: bottom 20 most routine tags (passive driving, in green).
  Both panels share the same x-axis for direct comparison.
  Tags were not seen during training.}
  \label{fig:correlation}
\end{figure}

For each qualifying tag we compute the mean surprise score over all
scenarios carrying that tag, then rank tags by mean score.
Figure~\ref{fig:correlation} shows the top 20 and bottom 20 tags.

The five highest-scoring tags are:

\begin{center}
\small
\begin{tabular}{lc}
\toprule
Tag & Mean surprise \\
\midrule
stopping\_at\_traffic\_light\_without\_lead & 2.008 \\
starting\_unprotected\_cross\_turn & 1.978 \\
starting\_left\_turn & 1.960 \\
starting\_protected\_cross\_turn & 1.891 \\
near\_pedestrian\_on\_crosswalk\_with\_ego & 1.834 \\
\bottomrule
\end{tabular}
\end{center}

\begin{flushleft}
The three lowest-scoring tags are \texttt{stationary\_in\_traffic}
(1.522), \texttt{near\_multiple\_pedestrians} (1.521), and
\texttt{following\_lane\_with\_lead} (1.406). The model assigns
higher scores to scenarios requiring active decision-making
and lower scores to passive or static situations, without
observing any of these labels during training.
\end{flushleft}

\FloatBarrier
\subsection{Ablation Studies}

We conduct four ablation experiments to establish that the tag
correlation result is not an artefact of the analysis pipeline,
the architecture, or a simple kinematic property of the data.
For all ablations we report Spearman rank correlation $\rho$
between the tag ordering produced by each method and the
tag ordering produced by the trained JEPA model, as well as
the score spread (max minus min of per-tag mean scores).

\textbf{Ablation 1: Shuffled scores.}
We randomly permute the surprise score list, breaking the
correspondence between scores and scenarios, and recompute
the per-tag mean scores. This tests whether the correlation
analysis pipeline can manufacture signal from noise.
The shuffled scores produce $\rho = -0.126$ ($p = 0.348$),
confirming the pipeline finds no pattern in unstructured noise.

\textbf{Ablation 2: Random encoder.}
We replace the trained model with a randomly initialised copy
and recompute all surprise scores. If any embedding geometry
sufficed to produce the result, a random encoder would
reproduce it. The random encoder produces $\rho = 0.024$
($p = 0.860$), showing that trained weights are necessary.

\textbf{Ablation 3: Constant-velocity baseline.}
We replace the neural network with a physics baseline that
extrapolates each agent's current velocity forward 2.5 seconds
and measures position prediction error as the surprise score.
The correlation between CV and JEPA tag orderings is moderate
($\rho = 0.314$, $p = 0.017$), but the CV score spread is 0.464
versus JEPA's 0.602, a 30\% reduction in discriminative power.
More critically, CV assigns \emph{low} scores to many
high-complexity scenarios: events such as unprotected turns
and pedestrian crossings begin with slow, controlled approach
behaviour. The constant-velocity predictor sees low kinematic
error during the approach and assigns low surprise. JEPA
captures the interaction dynamics and assigns high surprise
even before the critical event begins.

\begin{figure}[htbp]
  \centering
  \includegraphics[width=\textwidth]{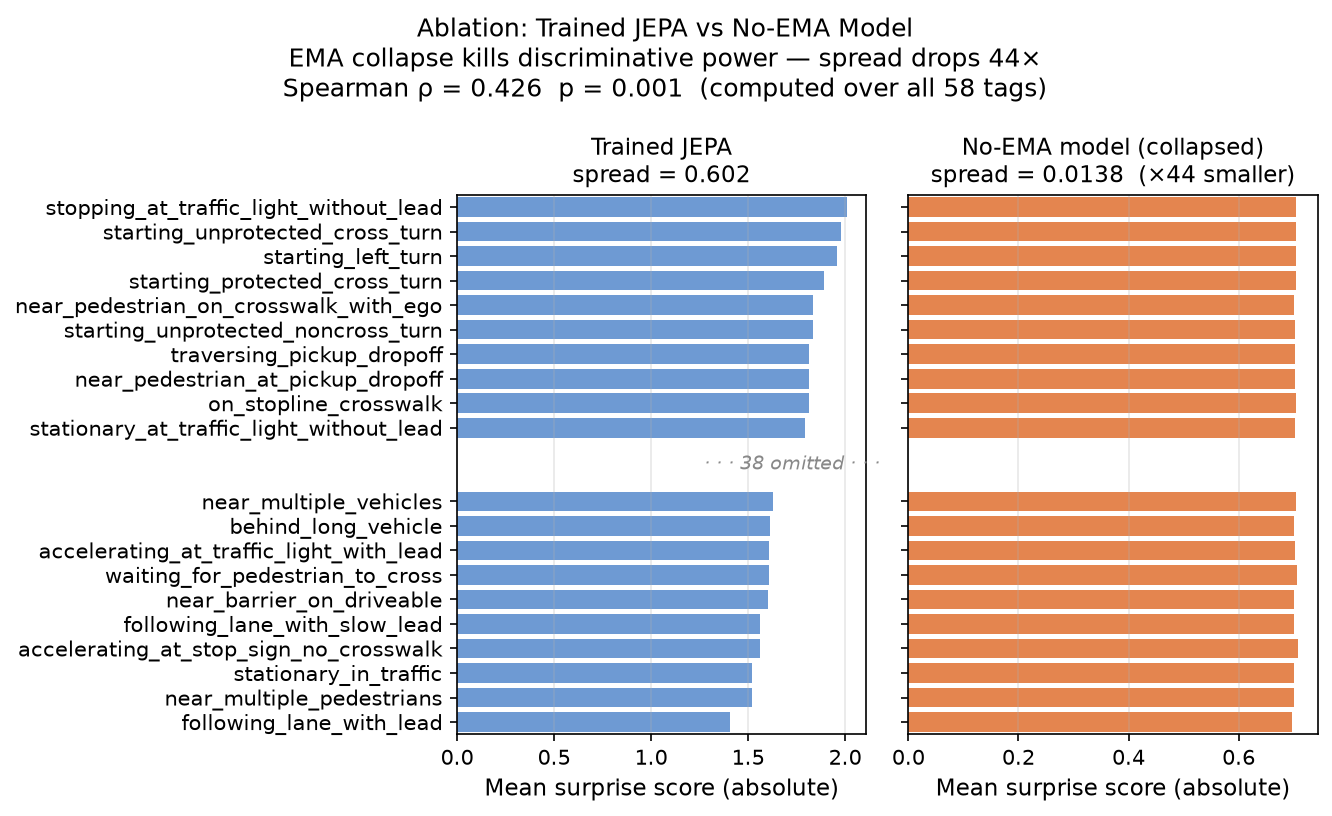}
  \caption{Ablation: per-tag surprise scores for the trained JEPA
  model (left) vs.\ no-EMA model (right); top 10 and bottom 10 tags
  shown (ranked by trained score). Without EMA the representation
  collapses: score spread drops from 0.602 to 0.014
  ($43{\times}$ smaller), making scenario discrimination
  impossible.}
  \label{fig:noema}
\end{figure}

\textbf{Ablation 4: No-EMA training.}
We retrain the model with EMA decay $\alpha = 0.0$, which
sets the target encoder equal to the context encoder at every
training step. This removes the mechanism that prevents
representation collapse.
Collapse is detected from epoch 3 onwards (cosine similarity
$> 0.998$, latent variance $< 0.002$). The resulting score
spread is 0.014 versus JEPA's 0.602, a 44-fold reduction.
All 1,322 scenarios receive nearly identical surprise scores
near the global mean, making discrimination impossible.
Figure~\ref{fig:noema} shows the collapse in score
distributions. This confirms that the EMA update is the
specific mechanism responsible for the learned signal.

\FloatBarrier
\subsection{Downstream Anomaly Detection}

We evaluate whether the surprise score is useful as a
zero-shot anomaly detector. We define 577 of the 1,322
scenarios (43.6\%) as ``complex'' based on the presence of
any of 18 tags covering unprotected turns, crosswalk
interactions, pedestrian proximity, and high lateral
acceleration. The model has no access to these labels.

We rank all scenarios by surprise score and measure
Precision@K: the fraction of the top-$K$ ranked scenarios
that are genuinely complex. We compare the trained JEPA model
against a constant-velocity baseline, the no-EMA model, and
a random encoder.

\begin{figure}[htbp]
  \centering
  \includegraphics[width=0.75\textwidth]{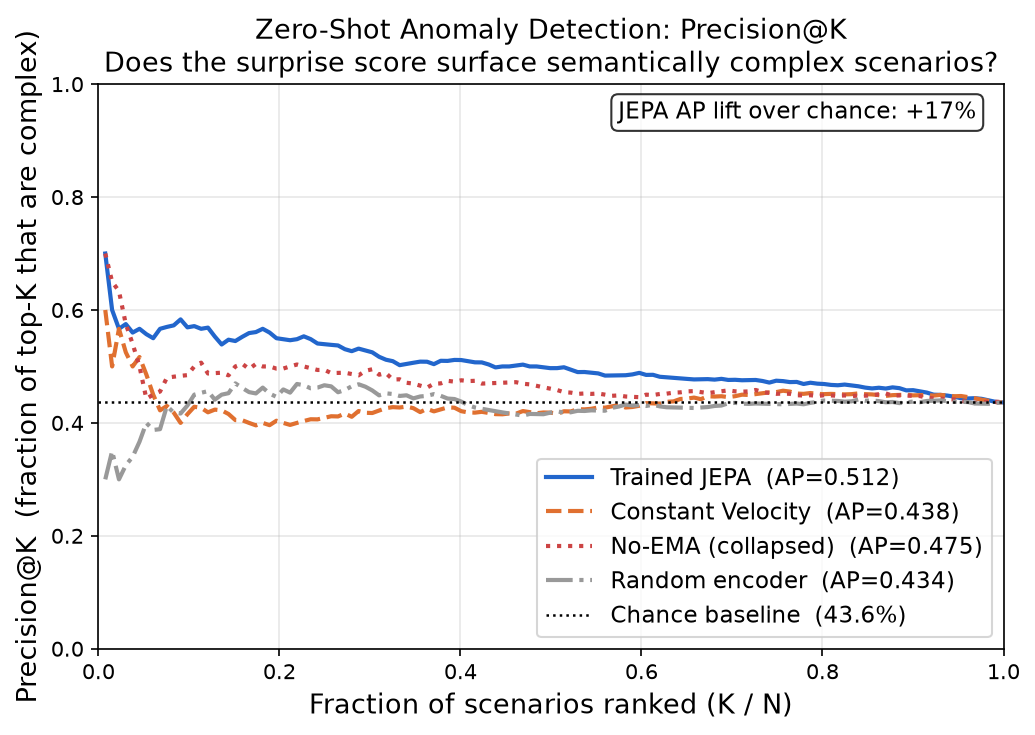}
  \caption{Precision@K curves for all four methods. The trained
  JEPA model (blue solid line) is the only one that stays above
  chance throughout the full ranking. The constant-velocity
  baseline (orange dashed) falls below chance after the top 5\%,
  as it assigns low surprise to slow-approach complex events.}
  \label{fig:topk}
\end{figure}

Table~\ref{tab:downstream} reports precision at four cutoffs
and Average Precision (AP). Figure~\ref{fig:topk} shows the
full precision-recall curve.

\begin{table}[htbp]
  \centering
  \small
  \caption{Precision@K for complex scenario detection.
  Base rate (chance) is 43.6\%.}
  \label{tab:downstream}
  \resizebox{\columnwidth}{!}{%
  \begin{tabular}{lccccr}
    \toprule
    Model & @5\% & @10\% & @20\% & @30\% & AP \\
    \midrule
    Trained JEPA & \textbf{56.1} & \textbf{57.6} & \textbf{55.3} & \textbf{52.8} & \textbf{0.512} \\
    Random encoder & 65.2 & 55.3 & 50.0 & 47.7 & 0.489 \\
    No-EMA & 47.0 & 49.2 & 49.6 & 49.2 & 0.475 \\
    Const. velocity & 50.0 & 40.9 & 39.8 & 41.7 & 0.438 \\
    Chance baseline & 43.6 & 43.6 & 43.6 & 43.6 & 0.436 \\
    \bottomrule
  \end{tabular}%
  }
\end{table}

The trained JEPA model is the only one that remains
consistently above the chance baseline across the full
ranking. The random encoder achieves high precision at very
small $K$ due to sampling variance, but decays toward chance
by $K = 20\%$. The constant-velocity baseline falls below
chance after the top 5\%, confirming that kinematic prediction
error alone is not a reliable complexity proxy.

\FloatBarrier
\subsection{Qualitative Results}

\begin{figure}[htbp]
  \centering
  \includegraphics[width=\textwidth]{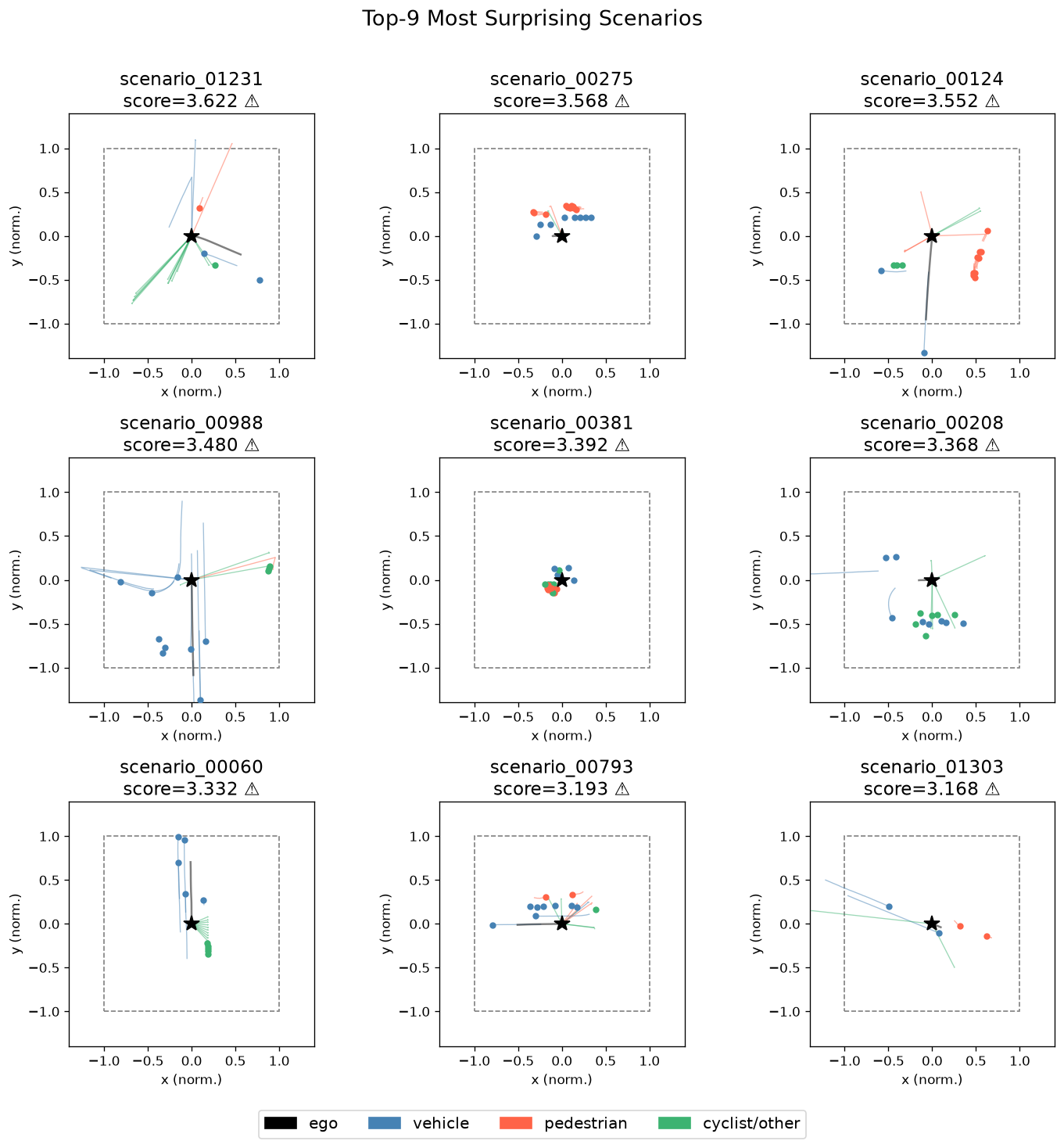}
  \caption{Top 9 most surprising scenarios ranked by JEPA surprise
  score. Each subplot shows agent trajectories over 5 seconds;
  ego vehicle in blue, other agents in grey. High-surprise scenes
  show dense agent clustering near intersections, turning
  manoeuvres, and pedestrian crossings.}
  \label{fig:birdseye_top}
\end{figure}

\begin{figure}[htbp]
  \centering
  \includegraphics[width=\textwidth]{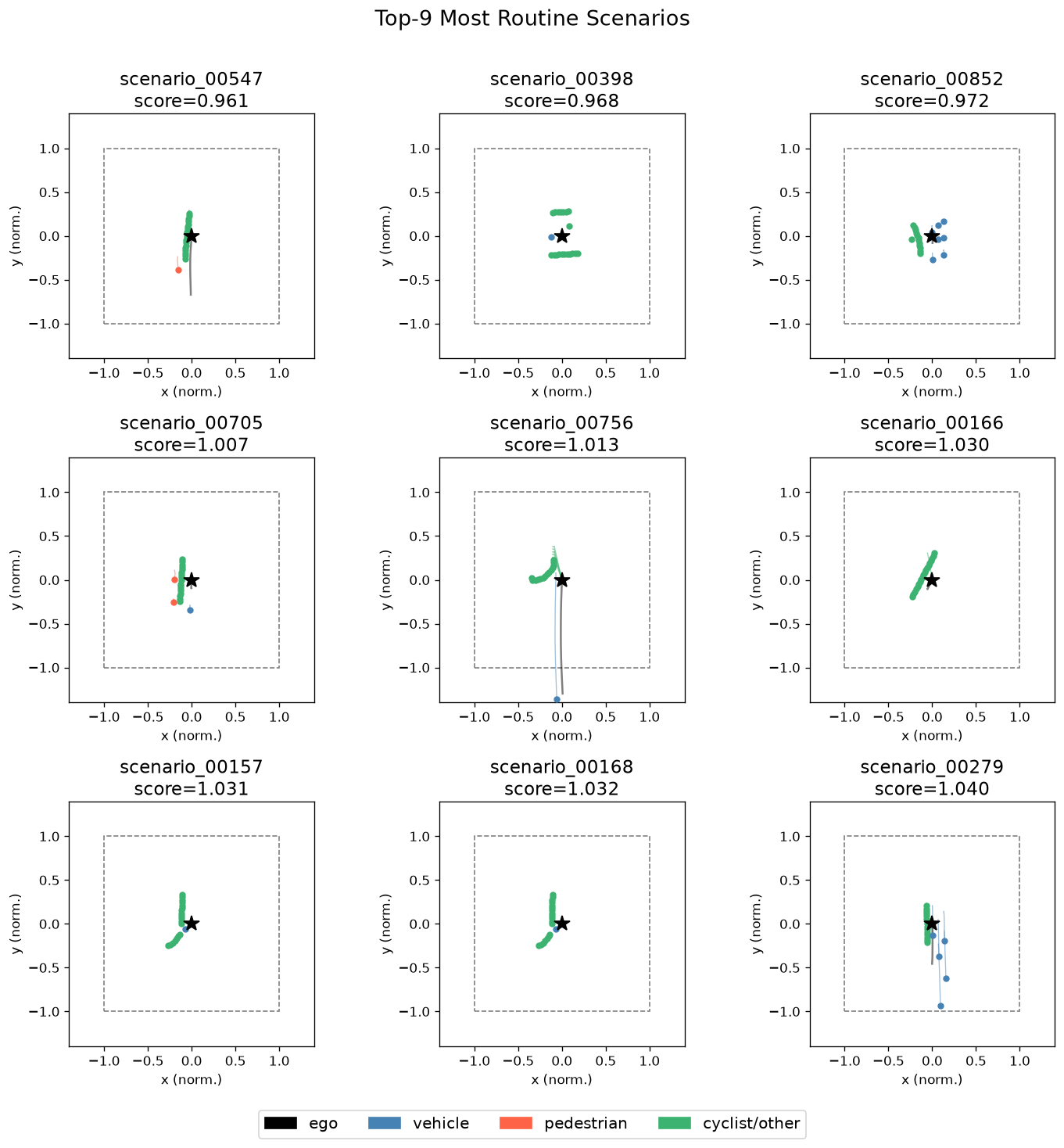}
  \caption{Bottom 9 most routine scenarios. Sparse agents moving
  in parallel lanes with little interaction; the model correctly
  assigns low surprise to these passive driving situations.}
  \label{fig:birdseye_bot}
\end{figure}

Figures~\ref{fig:birdseye_top} and~\ref{fig:birdseye_bot} show bird's-eye views of the nine
most surprising and nine most routine scenarios as ranked by
the surprise score. The high-surprise scenarios show dense
agent clustering near intersections, crossing pedestrians,
and multi-agent conflict points. The routine scenarios show
sparse agents moving in parallel lanes with little interaction.

% ─────────────────────────────────────────────────────────────────
\FloatBarrier
% ─────────────────────────────────────────────────────────────────
\section{Discussion}

\textbf{What the model learns.}
The surprise score reflects how well the model predicts
the compressed essence of the next 2.5 seconds. It fires on
scenarios with unusual agent interactions, not on visual
properties. This is a direct consequence of using structured
agent state data rather than pixels: there are no spurious
visual signals to latch on to.

\textbf{Why constant velocity underperforms.}
The constant-velocity baseline assigns low surprise to
scenarios that begin with controlled deceleration toward an
intersection. These are precisely the scenarios where
complex events occur next. The JEPA model, having seen
thousands of approach patterns, associates this deceleration
profile with upcoming complexity and assigns higher surprise.
This difference explains why CV falls below chance in the
downstream evaluation while JEPA does not.

\textbf{Why EMA matters.}
Without EMA, the target encoder is updated to match the
context encoder exactly at each step. This creates a
shortcut: the predictor can trivially minimise loss by
learning the identity function. EMA keeps the target
encoder as a slowly-moving historical average, which
closes the shortcut and forces the predictor to learn
genuine temporal structure. The 44-fold collapse in score
spread in Ablation 4 quantifies how much of the signal
depends on this mechanism.

\textbf{Limitations.}
This work uses the nuPlan mini subset, which contains
1,322 scenarios. The full nuPlan dataset has substantially
more data; we expect scores to become more calibrated
with scale. The current implementation is coupled to the
nuPlan data pipeline and would require a new tokenization
layer to work with other datasets. The downstream evaluation
has a relatively high base rate (43.6\%), which limits the
apparent precision lift. A more imbalanced evaluation set
would produce more striking precision numbers for the same
underlying model quality.

% ─────────────────────────────────────────────────────────────────
\section{Conclusion}

We trained a minimal JEPA-style world model on structured
driving agent data and showed that temporal prediction error
serves as a reliable zero-shot proxy for driving scenario
complexity. Without observing any labels, the model ranks
unprotected turns and crosswalk scenarios as more surprising
than lane-following and stationary traffic. Four ablation
experiments confirm that this signal depends on the trained
weights and specifically on the EMA mechanism that prevents
collapse. A downstream anomaly detection evaluation shows
consistent improvement over chance across the full scenario
ranking, with the constant-velocity kinematic baseline
providing a weaker and sometimes misleading signal.

The simplicity of the approach is a strength: a single model
trained end-to-end with a standard self-supervised objective
produces a useful complexity score at no annotation cost.
This makes JEPA-style world models a practical tool for
automatic dataset curation in autonomous driving.

% ─────────────────────────────────────────────────────────────────
\section*{Acknowledgements}

The nuPlan dataset is provided by Motional. Training was
conducted on an Apple M5 Max using the PyTorch MPS backend.

% ─────────────────────────────────────────────────────────────────
\FloatBarrier
\bibliographystyle{plainnat}
\bibliography{references}

\end{document}